\documentclass[a4]{article}
\usepackage{cite}
\usepackage{amsmath}
\usepackage{amssymb}
\usepackage{bbm}
\usepackage[linesnumbered,boxed]{algorithm2e}
\SetAlCapSkip{\baselineskip}

\SetCommentSty{SmallCommentFont}
\usepackage{textcomp}
\usepackage{graphicx,float}
\usepackage{ifpdf}
\usepackage{ifthen}
\ifpdf
\usepackage[bookmarksnumbered]{hyperref}
\else
\usepackage[bookmarks=false]{hyperref}
\fi
\SetKwInput{KwInput}{Input}
\SetKwInput{KwOutput}{Output}
\SetKwInput{KwInputOutput}{Input/Output}
\newcommand{\Mr}[1]{\mathrm{#1}}%
\newcommand{\Ms}[1]{\mathsf{#1}}%
\newcommand{\Ts}[1]{\textsf{#1}}%
\newcommand{\Tt}[1]{\texttt{#1}}%
\newcommand{\Tb}[1]{\textbf{#1}}%
\newcommand{\Em}[1]{\emph{#1}}
\newcommand{\Ul}[1]{\underline{#1}}
\newcommand{\Ol}[1]{\overline{#1}}
\newcommand{\Ba}{\begin{align}}%
\newcommand{\Bc}{\begin{center}}%
\newcommand{\Ec}{\end{center}}%
\newcommand{\Be}{\begin{enumerate}}%
\newcommand{\Ee}{\end{enumerate}}%
\newcommand{\Bi}{\begin{itemize}}%
\newcommand{\Ei}{\end{itemize}}%
\newcommand{\MbB}{\ensuremath{\mathbb{B}}}

\newcommand{\MbN}{\ensuremath{\mathbb{N}}}
\newcommand{\MbP}{\ensuremath{\mathbb{P}}}
\newcommand{\MbQ}{\ensuremath{\mathbb{Q}}}
\newcommand{\MbR}{\ensuremath{\mathbb{R}}}

\begin{document}
\newcommand{\nographics}{0}
\newcommand{\usegraphics}[5]
%
%
%
%
{\ifthenelse{\nographics = 1}
{\begin{center}\fbox{\fbox{\parbox[t]{0.25\textwidth}
{\textsf{\huge X}\begin{center}
\ifpdf{#5.pdf}\else{#5.eps}\fi
\end{center}}}}
\end{center}}
{\ifpdf
{\rotatebox{#2}{\resizebox{#3}{#4}{\includegraphics{#5.pdf}}}}
\else{\rotatebox{#1}{\resizebox{#3}{#4}{\includegraphics{#5.eps}}}}
\fi}}
\title{Fast Accurate Defect Detection in Wafer Fabrication}
\date{\today}
\author{Thomas Olschewski\footnote{thomas.olschewski@tu-dresden.de},
Technische Universit\"at Dresden}
\begin{titlepage}
\maketitle
\begin{abstract}
  A generic fast method for object classification is proposed.  In addition, a
  method for dimensional reduction is presented. The presented algorithms have
  been applied to real-world data from chip fabrication successfully to the
  task of predicting defect states of tens of thousands of chips of several
  products based on measurements or even just part of measurements.  Unlike
  typical neural networks with a large number of weights to optimize over, the
  presented algorithm tries optimizing only over a very small number of
  variables in order to increase chances to find a global optimum.  Our
  approach is interesting in that it is fast, led to good to very good
  performance with real-world wafer data, allows for short implementations and
  computes values which have a clear meaning easy to explain.
\end{abstract}
\end{titlepage}

\tableofcontents
\section{Acknowledgements}
\label{sec:Acknowledgements}
I thank Zolt\'{a}n Sasv\'{a}ri for providing the environment which made this
research possible, for carefully reading the manuscript and for some
corrections.
\\~\\
Research leading to these results has received funding from the iRel40
project.  iRel40 is a European co-funded innovation project that has been
granted by the \Tb{ECSEL} Joint Undertaking (JU) under grant agreement
\Tb{n$^\mathbf{o}$ 876659}. The funding of the project comes from the Horizon
2020 research programme and participating countries. National funding is
provided by Germany, including the Free States of Saxony and Thuringia,
Austria, Belgium, Finland, France, Italy, the Netherlands, Slovakia, Spain,
Sweden, and Turkey.
\\~\\
\Tb{Disclaimer:} The document reflects only the author’s view and the JU is
not responsible for any use that may be made of the information it contains.
%
%
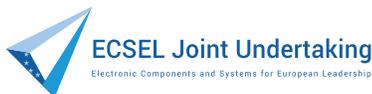
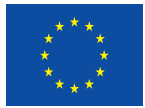
\begin{figure}[H]
\usegraphics{270}{0}{0.4\textwidth}{0.1\textwidth}{logo-ECSEL-1}
\qquad\qquad
\usegraphics{270}{0}{0.15\textwidth}{0.1125\textwidth}{flag-yellow-cropped}
\end{figure}
%
\section{Introduction}
\label{sec:Introduction}
In this paper we propose a generic fast method for object classification some
specialisations of which serve to detect anomalies in measurement data of
wafer fabrication.  The basic task of this application is predicting the
positive/negative state of objects knowing only measurement data.  In the
wafer data application this translates to predicting from measurement data
early in the fabrication whether some chip will turn out to be
defective---i.e. is a positive object or not---later.

Our approach is interesting in that it is fast, led to good to very good
performance with real-world wafer data, allows for short implementations and
computes values which have a clear meaning easy to explain.

The baselying principles are using thresholdings of coordinate-wise normalized
data in order to derive bit patterns from the input data, computing
similarities to a small set of training objects known to be positive and
finding an optimal cutoff in order to classify objects yet unknown.  We
propose a selection of thresholdings and similarity measures which have been
proven successful in the wafer data application.  In its basic form there is
one free parameter $t$ for which good value selections can be obtained easily.

We show how this method for object classification follows from a logical
algorithm tackling a somewhat easier problem, the problem of object
identification, i.e. the problem of deciding whether some object belongs to a
known set of (training) objects.

We point out that in several practical settings our algorithms allow for
identifying \emph{prototypical defect chips} which in our nomenclature are
chips being especially important in the sense that they possess properties
which allow for accurately classifying lots of other, yet unclassified, chips.

Furthermore we propose a fast method for reducing input data dimensionality
which serves as an optional preprocessing step.

Time and space complexities of our algorithms are quasi-linear and thus are
good-natured enough in order to allow for classifying large sets of objects
with high-dimensional input data on normal PC hardware.  In our applications
we could classify e.g. 11328 chips in 915 dimensions with accuracy 99.8\% and
kappa 0.979 within a few seconds.
\\~\\
The algorithm we describe here possesses several advantages over typical usage
of neural nets for machine learning by back-propagation\cite{Rumelhart:1986}.
Firstly, in our applications there is only a very limited number of unknowns
to be optimized over, namely the parameter $t$ of subsection
\ref{ssec:SettingsOfPractical} being used in thresholding the input data and
thus implicitly contained in Algorithm 1, and the cut-off $C$ in Algorithm 3
which converts the guesses into a binary positive/negative prediction.

So, unlike with back-propagation we are not faced with thousands or even
millions of unknown weights spanning a giant search range to be optimized over
but with only one or two which enables approximating the global optimum for
arbitrary measures of prediction quality.  With our methods, it is not
hopeless to try finding a global optimum in reasonable time on ordinary PC
hardware as the graphs of target functions are rather good-natured having few
local optima.  Secondly, the algorithmic meaning of the few unknowns in our
algorithms is easy to understand which we see as a step towards XAI
(explainable AI), see for example \cite{Arrieta:2020}.


The goals of a series of ongoing research projects for improving quality
control in chip fabrication suggest the following properties for an ideal
algorithm in order to be economically useful:

\Be
\item Not requiring Gaussian distribution of single---or several or
  all---parameters.
\item High true-positive-false-positive quotients (TP/FP):
  As few as possible good devices shall be scrapped for sorting out
  bad devices.
\item High efficiency: Data should be processed in fabrication real--time.
\item Ability of coping with large amount of data: One typical lot
  $\approx 30-100$ MiB of data.
\item Reducing the number of tests: Measurements are costly and time consuming
  to different extents so it is desirable to need less of them without
  reducing the quality of defect prediction.
\Ee
Algorithms 3 and 4 which will be presented in subsection \ref{ssec:Properties}
meet all of these goals.
\subsection{Overview}
\label{ssec:Overview}
The rest of this paper is organized as follows.  In section
\ref{sec:TheAlgorithms} we will describe our algorithms and two applications.
Section \ref{sec:ExperimentalSetup} is about the experimental setup and the
task to be solved.  In section \ref{sec:Results} we will evaluate and analyse
the results we obtained from a larger number of test runs and describe a
method for dimensional reduction applied sucessfully. In section
\ref{sec:Conclusion} we will add concluding thoughts.

\section{The Algorithms}
\label{sec:TheAlgorithms}
\subsection{Definitions}
\label{ssec:Definitions}
Let $B_{i}\subseteqq B$, $i=1,\ldots,s$ be subsets of a fixed base set $B$
with characteristic functions $\mathbbm{1}_{B_{i}}$. We define
$P_{i}=\mathbbm{1}_{B_{i}}$ for shortness and $\MbP=(P_{1},\ldots,P_{s})$.
For every $x\in B$ define $I(x)=\{i\mid x\in B_{i}\}$. For every pair $x,y\in
B$ we define
\[ s(x,y)=
    \begin{cases}
    \frac{|I(x)\cap I(y)|}{|I(x)|} & I(x)\neq \O \\
    0 & I(x) = \O
   \end{cases}
\]
and call it \Em{coincidence index} of $x$ and $y$.  Clearly, $0\leqq
s(x,y)\leqq 1$ where $0$ means disjointness of $I(x)$ and $I(y)$ and $1$ means
$I(x)\subseteqq I(y)$.

Let $D\subseteqq B$ be a set henceforth called \Em{data points} and
$T\subseteqq D$ a set henceforth called \Em{training set}.

In what follows we will use the following abbreviation:
\begin{align*}
\MbB &= \{0,1\}
\end{align*}
\subsection{Dense Formulation}
\label{ssec:DenseFormulation}
Alternatively, the formula for the coincidence index can be rewritten by
using incidence vectors as follows:
\[ s(x,y) = \frac{\sum_{i=1}^{s}P_{i}(x)\cdot P_{i}(y)}
                 {\sum_{i=1}^{s}P_{i}(x)}\qquad(x,y\in B)
\]
We set $s(x,y)=0$ in case $\sum_{i=1}^{s}P_{i}(x)=0$.
We call
\[ v(x) = \big(P_{1}(x),\ldots,P_{s}(x)\big)\in\mathbb{B}^{s}
\]
the \Em{incidence vector} of $x$ relative to $\MbP$.  Then we can write
\[ s(x,y) = \frac{v(x)\cdot v(y)}
                 {v(x)\cdot v(x)}\qquad(x,y\in B)
\]
with $a\cdot b$ being the inner product, computed in \MbN, of two 0-1 vectors
$a,b$.

We can rewrite this further as
\[ s(x,y) = \frac{H\big(v(x)\circ v(y)\big)}
                 {H\big(v(x)\big)}\qquad(x,y\in B)
\]
where $H(v)$ is the Hamming weight 
\[ H(v)=\sum_{i=1}^{s}v_{i}
\] 
of vector $v\in\MbB^{s}$ ($H(v)\in\{0,\ldots,s\}$) and $v\circ w=(v_{1}\cdot
w_{1},\ldots,v_{s}\cdot w_{s})$ shall denote the component-wise multiplication
of two vectors $v$ and $w$.  If limited to 0-1 vectors $v$ and $w$, $v\circ
w=(v_{1}\wedge w_{1},\ldots,v_{s}\wedge w_{s})$ and
\[  H\big(v\circ w\big)=\big|\{i\mid v_{i}=w_{i}=1\}\big|.
\]
Now we consider the following problem.
\\\mbox{}\\
\Tb{Given:} A base set $B$, data set $D\subseteqq B$, training set
$T\subseteqq B$, sets $B_{i}\subseteqq B$, $i=1,\ldots,s$, and a data point
$x\in D$. 
\\\mbox{}\\
\Tb{Task:} Compute 
\[ s(x, T) = \max\{s(x,y)\mid y\in T\}.
\]
We call $s(x, T)$ the \Em{coincidence index of $x$ relative to $T$}.  Since
$s(x, T)=1$ trivially if $x\in T$ we will consider only data points $x\not\in
T$ henceforth.

\Tb{Computability assumptions.}  The predicates $P_{i}:B\rightarrow\MbB$ must
be computable in the machine model used, elements of $B$ must be representable
in that machine model. See, for example, \cite{Savage:1998} for an
introduction to these topics.

\Tb{Examples.}  Machine model = Turing machine and $B=\MbQ^{n}$. $B_{i}$ must
be Turing decidable sets in this case.
\subsubsection{Extensions}
\label{sssec:extensions}
$s(x,y)$ is not limited to the coincidence index as defined in subsection
\ref{ssec:Definitions}.  Basically, if
$f:\MbB^{s}\times\MbB^{s}\longrightarrow[0,1]\subseteqq\MbQ$ is any function
expressing a degree of similarity of two 0-1 vectors in some sense then we can
set
\[ s(x,y) = f\big(v(x),v(y)\big).
\]
Cohen's kappa function is another example for $f$ we used in practice, see
subsection \ref{sssec:CutoffSelection} for a definition.

In addition, $s(x,T)$ can be substituted by other functions which promise to
indicate a spread in between positive and negative objects.  For example, we
investigated $s(x,T)=\min\{s(x,y)\mid y\in T\}$, too.
\subsection{$s(\cdot,y)$ as a Linear Functional}
\label{ssec:AsALinearFunctional}
With $x$---and thus $v(x)$---being fixed, $s(x, y)$ can be written as a linear
functional in the vector $v(y)$.  For, by setting
\[ L_{x}(a) =  \frac{v(x)\cdot a}{H\big(v(x)\big)}
\]
we can write
\[  s(x, y) = L_{x}\big(v(y)\big)
\]
where $v\cdot w$ is the euclidian inner product and $H(\cdot)$ is the Hamming
weight function.

Using the formula for ``cosine similarity'' of 0-1 vectors $v(x),v(y)$
\[ C\big(v(x),v(y)\big) = \frac{v(x)\cdot v(y)}{\|v(x)\|\cdot\|v(y)\|}
\]
and using 
\[ \|v\| = H(v)^{1/2}
\]
for 0-1 vectors $v$ with the euclidian distance $\|\cdot\|$ we can write
\begin{align*} 
    s(x,y) 
    &= L_{x}\big(v(y)\big) = \frac{v(x)\cdot v(y)}{H\big(v(x)\big)} \\
    &= \frac{v(x)\cdot v(y)}
            {{H\big(v(x)\big)}^{1/2}\cdot{H\big(v(y)\big)}^{1/2}} 
       \cdot \frac{{H\big(v(y)\big)}^{1/2}}{{H\big(v(x)\big)}^{1/2}} \\
    &= C\big(v(x),v(y)\big)
       \cdot \frac{{H\big(v(y)\big)}^{1/2}}{{H\big(v(x)\big)}^{1/2}}
\end{align*}
So $s(x,y)$ is the ``cosine similarity'' of 0-1 vectors $v(x),v(y)$,
multiplied by
\[ q(x,y)=\frac{{H\big(v(y)\big)}^{1/2}}{{H\big(v(x)\big)}^{1/2}}
\]
\subsection{Machine Learning}
\label{ssec:MachineLearning}
We extend the above setup by a function
\[ p:D\rightarrow\MbB
\]
which assigns one bit $p(x)$ to every data point $x\in D$.  We call $x$ a
\Em{positive} data point if $p(x)=1$ and a \Em{negative} data point if
$p(x)=0$.  Let
\[ D^{1}=\{x\in D\mid p(x)=1\},\quad D^{0}=\{x\in D\mid p(x)=0\}
\]
We limit ourselves to considering exclusively positive data points for being
used as training data, i.e. in what follows we presuppose
\[ T\subseteqq D^{1}
\]
\subsubsection{The General Algorithm}
\label{sssec:TheGeneralAlgorithm}
Let $B\subseteqq\MbR^{n}$ and $D\subseteqq B$ be a set of data points.
\\\mbox{}\\
\begin{algorithm}[H]
\KwInput{$B=\MbR^{n}$, $D\subseteqq B$, $p\in[0,100]$}
\caption{General algorithm}
scale($D$)\\
$T=\mathrm{select(p, D^{1})}$\\
\If{$T=\text{\O}$}
{
    stop
}
\For{$x\in D\setminus T$}
{
  compute s($x$, $T$)
}
$Av^{0} := \mathrm{avg}\{ s(x,T)\mid x\in D^{0}$\}\\
$Av^{1} := \mathrm{avg}\{ s(x,T)\mid x\in D^{1}$\}
\end{algorithm}
~\\
$\mathrm{select(p, M)}$ is a function which draws 
\[
    \lceil |M|\cdot\frac{p}{100}\rceil
\]
elements from a set $M$ randomly without repetition and $s(x, T)$ is as
described above.  Here---and in other places---we neglect the trivial case of
$T$ being empty. scale($D$) refers to the following function.
\subsubsection{Scaling}
\label{sssec:Scaling}
\begin{algorithm}[H]
\KwInputOutput{$D=\{x_{1},\ldots,x_{m}\}\subseteqq \MbR^{n}$}
\caption{scale($D$)}
\For{$j=1\ldots n$}
{
  compute $\mu_{j},\sigma_{j}$\\
  \tcp{mean and standard deviation of the $j$-th components
       of $x_{1},\ldots,x_{m}$}
}
\For{$i=1\ldots m$}
{
  \For{$j=1\ldots n$}
  {
    $x_{i,j} := \begin{cases}
               \frac{x_{i,j}-\mu_{j}}{\sigma_{j}},
                   & \sigma_{j}\neq0\\
               0, & \sigma_{j}=0
               \end{cases}$
  }
}
\end{algorithm}
~\\
$D$ has mean 0 and standard deviation 1 after scaling.  After scaling $D$ the
general algorithm does not operate on the data points
$x_{i}=(x_{i,1},\ldots,x_{i,n})$ actually given but on the vectors
that quantify how far (by how many standard deviations $\sigma_{j}$)
$x_{i,j}$ is away from the mean $\mu_{j}$.
\subsubsection{Computing Predictions}
\label{sssec:ComputingPredictions}
Let $B$, $D$, $p$ and $T$ the same as in Algorithm 1.
\\\mbox{}\\
\begin{algorithm}[H]
\KwInput{$s(x,T)$ as computed by Algorithm 1, $C\geqq0$}
\caption{Computing predictions}
\For{$x\in D\setminus T$}
{
    $F(x, T, C) := \begin{cases}1, & s(x,T)\geqq C \\
                                0, & s(x,T) < C
                   \end{cases}$
}
\end{algorithm}
~\\
$C$ operates as a cutoff for digitizing $s(x,T)$.  There is a wide range of
methods which can be used for specifying $C$ of which we describe just two.
\subsubsection{Cutoff Selection}
\label{sssec:CutoffSelection}
In what follows, we assume $Av^{1}-Av^{0}\geqq0$\footnote{If this does not
  hold true in a specific application then one can replace $F(x, T, C)$ by its
  logical inverse $\neg F(x, T, C) = 1-F(x, T, C)$ in order to achieve
  $Av^{1}-Av^{0}\geqq0$.  We would interprete this as an indication of an
  insufficient training set $T$---it may be too small, containing too few
  characteristic samples, or samples not characteristic enough--- or that this
  algorithm is not apt to this specific input data $B$, $D$, $p$ and the set
  of predicates $\MbP$.} for $Av^{1},Av^{0}$ as defined in Algorithm 1.
\\\mbox{}\\
\Tb{Na\"ive cutoff selection}.
\\\mbox{}\\
\[
    C := \frac{Av^{1}-Av^{0}}{2}
\]
This is a better-than-nothing selection which can be computed very fast but
may deliever predictions $F(x,T,C)$ far from what can be achieved with a more
sophisticated cutoff.
\\\mbox{}\\
\Tb{Optimizing cutoff for some statistical quantity $Q(v,w)$}.
\\\mbox{}\\
Let $D\setminus T = x^{1},\ldots,x^{m^{\ast}}$ an enumeration in some
fixed order and
\[
    \Ms{S} = \big(p(x^{1}),\ldots,p(x^{m^{\ast}})\big)
\]
the 0-1 vector of their positive/negative states.  Let be
\[
    \Ms{F}(C) = \big(F(x^{1},T,C),\ldots,F(x^{m^{\ast}},T,C)\big)
\]
the 0-1 vector of all $F(x,T,C)$ for $x\in D\setminus T$ in the same ordering
as in $\Ms{S}$.  Furthermore let be $Q(v,w)$ some statistical quantity which
represents some measure of similarity of two 0-1 vectors $v$ and $w$, such like
accuracy\footnote{Accuracy is the amount of coincident bits:
  $\mathrm{accu}(v,w)=\frac{1}{n}\cdot\sum_{i={1}}^{n}\big(v_{i}\cdot
  w_{i}+(1-v_{i})(1-w_{i})\big)$.}  or Cohen's kappa\footnote{Kappa is used
  for measuring agreement of two binary raters $v$ and $w$
  here\cite{Cohen:1960}.  For binary raters,
  $\mathrm{kappa}(v,w)=\frac{\mathrm{accu}(v,w)-p_{e}}{1-p_{e}}$ where
  $p_{e}=\frac{1}{n^{2}}(n_{v_{i}=0}\cdot n_{w_{i}=0} + n_{v_{i}=1}\cdot
  n_{w_{i}=1})$.}
\\\mbox{}\\
\begin{algorithm}[H]
\KwInput{$B,D,T,p$ as in Algorithm 1}
\caption{Optimize cutoff $C$}
\For{$C$ on a grid $\mathcal{C}\subset[0,1]$}
{
  Compute $\Ms{F}(C)$ by Algorithm 3\\
  Compute $Q\big(\Ms{F}(C),\Ms{S}\big)$
}
$Q_{opt} := \max\{Q\big(\Ms{F}(C),\Ms{S}\big)\mid C\in\mathcal{C}\}$\\
$C_{opt}:=\text{one value of $C$ where $Q_{opt}$ is reached}$
\end{algorithm}
~\\
For improvement in both complexity and controlling the quality of
approximation it is desirable to replace this brute-force type algorithm for
searching the optimum $(Q_{opt},C_{opt})$ by more refined methods.  In our
applications, it turned out that the interpolation from $\mathcal{C}$ to all
of $[0,1]\subset\MbR$ of the graph $\Gamma_{Q}$ of the function
$C\longrightarrow Q\big(\Ms{F}(C),\Ms{S}\big)$ is rather good-natured as for
finding global optima.  In the vast majority of settings $\Gamma_{Q}$ was a
hill-formed curve, i.e. increasing (decreasing) to the left (right) of
$(C_{opt},Q_{opt})$ the top of which was the global maximum over all $[0,1]$.
See Figure \ref{fig:kappa-cutoff-default} and Figure
\ref{fig:kappa-cutoff-abs} for two typical $\Gamma_{Q}$ examples.
\subsubsection{Refined Method For Cutoff Selection}
\label{sssec:RefinedMethodForCutoffSelection}
The brute-force method of Algorithm 4 has two disadvantages:
\Bi
\item
  Obtaining a good approximation of the global optimum takes many grid points to
  be evaluated which increases computation time.
\item
  The quality of approximation is bound to the step-width of the grid.
\Ei
Now we propose an algorithm which combines the advantage of the grid-based,
fixed-step-width Algorithm 4---namely, independency of the starting
point---with the advantages of flexible-step-width algorithms---improved speed
and better control over the quality of approximation.
\\~\\
\\\mbox{}\\
\begin{algorithm}[H]
\KwInput{$B,D,T,p$ as in Algorithm 1.
    $n_{steps}\in\MbN_{\geqq1}$, $\varepsilon>0$}
\caption{Optimize cutoff $C$, refined method}
Grid $\mathcal{C} :=\{t_{1},\ldots,t_{n_{steps}}\}$ equidistant in $[0,1]$\\
\For{$C$ on $\mathcal{C}\subset[0,1]$}
{
  Compute $\Ms{F}(C)$ by Algorithm 3\\
  Compute $Q\big(\Ms{F}(C),\Ms{S}\big)$
}
$Q_{opt,\mathcal{C}} := \max\{Q\big(\Ms{F}(C),\Ms{S}\big)\mid C\in\mathcal{C}\}$\\
$C_{opt,\mathcal{C}}:=\text{one value of $C$ where
      $Q_{opt,\mathcal{C}}$ is reached}$\\
$t_{a} := \max(0,C_{opt,\mathcal{C}}-\frac{1}{2n_{steps}})$\\
$t_{b} := \min(1,C_{opt,\mathcal{C}}+\frac{1}{2n_{steps}})$\\
$sw := \frac{t_{b}-t_{a}}{100}$ \qquad\qquad\qquad\qquad\tcp{step-width}
$P_{0} =(t_{0},Q_{0}) := \Big(t_{a},Q\big(\Ms{F}(t_{a}),\Ms{S}\big)\Big)$\\
$P_{1} =(t_{1},Q_{1}) := \Big(t_{a}+sw,Q\big(\Ms{F}(t_{a}+sw),
                       \Ms{S}\big)\Big)$ \\
$i := 1$\\
\While{true}
{
  \If{$|Q_{i}-Q_{i-1}|<\varepsilon$}
  {
    \Tb{break}
  }
  \If{$Q_{i}<0$}
  {
    $sw := |sw|\cdot 2$
  }
  \If{$Q_{i}>1$}
  {
    $sw := -|sw|\cdot 2$
  }
  \If{$Q_{i}\geqq Q_{i-1}$}
  {
    $sw := 1.5\cdot sw$
  }
  \If{$Q_{i}<Q_{i-1}$}
  {
    $sw := -\frac{sw}{2}$
  }
  $P_{i+1} =(t_{i+1},Q_{i+1})
       := \Big(t_{i}+sw,Q\big(\Ms{F}(t_{i}+sw),\Ms{S}\big)\Big)$\\
  $i=i+1$
}
$Q_{opt} := Q_{i}$\\
$C_{opt} := t_{i}$
\end{algorithm}
~\\
In this algorithm we search for the optimum cutoff $C_{opt,\mathcal{C}}$ on
the points of a grid with fixed-step-width $\frac{1}{n_{steps}}$ first in
order to get close to the global optimum.  In the \Tb{while} $true$ loop
following we examine the interval $[t_{a},t_{b}]$ of diameter
$\frac{1}{n_{steps}}$ centered at $C_{opt,\mathcal{C}}$ and initialize
the---from now on---flexible step-width $sw$ by, say,
$\frac{t_{b}-t_{a}}{100}$. Then we increase $t$ by $sw$ if the target function
$Q$ was non-decreasing when going from $t_{i-1}$ to $t_{i}$. We step by
$\frac{|sw|}{2}$ in the opposite direction if the target function $Q$ was
decreasing, supposing that we have overrun the peak of $Q$.  We break if the
most recent change of $Q$ has not exceeded $\varepsilon$.

Using this algorithm we can dramatically reduce $n_{steps}$ and thus save
computation time spent for searching on the fixed-step-width grid
$\mathcal{C}$ because it suffices to get to $[t_{a},t_{b}]$ containing the
global optimum now and approximate further by iterating in the \Tb{while}
$true$ loop. Because searching on $\mathcal{C}$ consists in a loop of
completely independent evaluations of $Q$ reducing $n_{steps}$ to
$\lfloor\frac{n_{steps}}{f}\rfloor$ improves computation time for searching on
$\mathcal{C}$ roughly by factor $f$.
\subsection{Settings of Practical Applications}
\label{ssec:SettingsOfPractical}
Let $B=\MbQ^{n}$, $B_{i}=\{x\in B\mid P_{i}(x)=1\}$ where
$P_{i}:B\longrightarrow\MbB$ might be any Turing computable functions.  The
following types of function are prototypical examples of $P_{i}$.  Let be
$t>0$ some constant.
\[ P^{e}_{i}(x)=\begin{cases}
    1, & x_{i}-t > 0 \\
    0, &\text{else}
  \end{cases}
\]
We will call $(i,x_{i})$ with $P^{e}_{i}(x)=1$ a \Em{$t$-excess} henceforth.
\[ P^{a}_{i}(x)=\begin{cases}
    1, & |x_{i}|-t > 0 \\
    0, &\text{else}
  \end{cases}
\]
We will call $(i,x_{i})$ with $P^{a}_{i}(x)=1$ an \Em{abs-$t$-excess}
henceforth.  Now let be $\Ul{R}_{i} \leqq \Ol{R}_{i}$ numerical constants.
\[ P^{R}_{i}(x)=\begin{cases}
    1, & x_{i} < \Ul{R}_{i}\quad\text{or}\quad x_{i} > \Ol{R}_{i}\\
    0, &\text{else}
  \end{cases}
\]
We will call $(i,x_{i})$ with $P^{a}_{i}(x)=1$ a \Em{ref-excess} hencefort.
In practice, $\Ul{R}_{i}$ and $\Ol{R}_{i}$ might be the lower and upper edge,
respectively, of a reference range.

Besides the above types of predicates $P_{i}$ which are related directly to
practical applications one can imagine more theoretical types of $B_{i}$ (and
thus $P_{i}$), e.g. linear half spaces ($L(x)>c$) or semi-algebraic sets
($p(x)>0$).
\subsection{Application 1: Identification of Positive Objects}
\label{ssec:Application1}
Let us set $B=\MbQ^{n}$ and assume we are given a data set $D\subseteq B$ and
a disjunctive subdivision $D=D^{1}\cup D^{0}$ into positive and negative data
points as in subsection \ref{ssec:MachineLearning}.  Set $T=D^{1}$ and $p=100$
This means we are using the set of \Em{all} positive data points as training
set.  In addition, we set cutoff $C=1$.

With this special settings Algorithm 3 can be used as a means to identify
positive data points.  Setting $C=1$ means that $F(x,T,C)$ as computed by
Algorithm 3 predicts $1$ (for ``positive object'') if and only if $s(x,T)=1$.
$s(x,T)=1$ if and only if $\exists y\in T$ with $s(x,y)=1$ by the very
definition of $s(x,T)$ in subsection \ref{ssec:DenseFormulation}.  By
definition of $s(x,y)$ in subsection \ref{ssec:DenseFormulation} this means
\[ 1 = \frac{\sum_{i=1}^{s}P_{i}(x)\cdot P_{i}(y)}
            {\sum_{i=1}^{s}P_{i}(x)}
\]
or
\[ \sum_{i=1}^{s}P_{i}(x)\cdot P_{i}(y) = \sum_{i=1}^{s}P_{i}(x)
\]
which is equivalent to:
\[ \forall i=1,\ldots,s:\big[P_{i}(x)=1\longrightarrow P_{i}(y)=1\big]
\]
So in this setting Algorithm 3 predicts an object $x$ under test to be a
positive object if there is at least one positive object $y$ in $T=D^{1}$ such
that all ``excesses'' $(i,x_{i})$ of $x$ are ``excesses'' of $y$, too.  If
this is the case for the object under test $x$ then we predict $x$ to be a
positive object and assign $y$ to it.
\subsubsection{Building and Solving a System of Implications}
\label{sssec:BuildingAndSolving}
The method for object identification described above embodies a special case
of Algorithm 3 by using the special settings of $T=D^{1},p=100,C=1$.  As an
alternative one can take a different route to the same goal by first building
a system of logical implications from all positive objects $y\in D^{1}$ (here
$T=D^{1}$)
\[ \exists y\in D^{1}:\forall i=1,\ldots,s:\big(P_{i}(x)=1\longrightarrow
  P_{i}(y)=1\big)
\]
The implication system defined by the ``excesses'' of all $y\in D^{1}$ can be
formulated as a Prolog program as well as the request whether a given object
$x\in D$ is a positive one.  The Prolog program finds those $y$ satisfying the
implication system by unification if there are any. We used
\Ts{GNU Prolog}\cite{AbreuCodognetDiaz:2012} for solving these implication
systems. See \cite{Kowalski:1979} for an introduction to this
topic.

At the end of this paragraph a Prolog program describing an implication system
of the type mentioned above is listed.  Each line starting with
\verb=defect(X) :- = is a rule which describes the thresholding of
measurements of one positive object whereas the line \verb=is_high(4, z)=
formalizes the fact that measurement of parameter 4 of the object to be
classified exceeded the threshold.  \Bc
\begin{minipage}[c]{0.95\textwidth}
\begin{small}
\begin{verbatim}
defect(X) :- is_high(16,  X) , is_high(19, X) , is_high(21,  X) , 
             is_high(22,  X) ,      [...]     , is_high(380, X) , 
             is_high(383, X). % [Line 7]
defect(X) :- is_high(3, X) , is_high(13, X) , [...] , 
             is_high(385, X). % [Line 15]
%
% [...]
%
defect(X) :- is_high(5,   X) , is_high(7,   X) ,      [...]      , 
             is_high(375, X) , is_high(378, X) , is_high(381, X) , 
             is_high(384, X). % [Line 816]

is_high(4,   z).
is_high(19,  z).
is_high(26,  z).
%
% [...]
%
is_high(319, z).
is_high(322, z).
\end{verbatim}
\end{small}
\end{minipage}
\Ec
\subsection{Application 2: Predicting Positive Objects}
\label{ssec:Application2}
In order to progress from object identification to predicting the
positive/negative state of some object under test $x$ yet unknown we do not try
to find an object $y$ of the training set whose set of ``excesses'' contains
the set of ``excesses'' of the object under test but we are going to determine
the degree of satisfiability by using the formula of subsection
\ref{ssec:DenseFormulation}:
\[ s(x, T) = \max_{y\in T}
            \frac{\sum_{i=1}^{s}P_{i}(x)\cdot P_{i}(y)}
                 {\sum_{i=1}^{s}P_{i}(x)}
\]
\begin{figure}
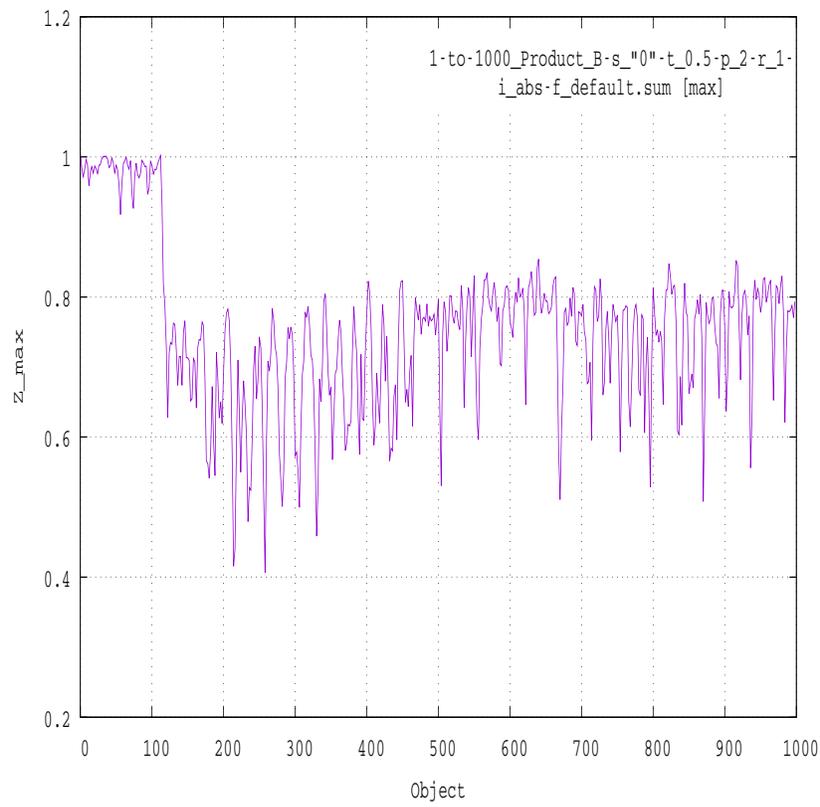

  \caption{Classifying 1000 chips with 2\% training set size.}
  \label{fig:class1000}
  \centering
  \usegraphics{270}{0}{0.9\textwidth}{0.9\textwidth}{1-to-1000-B-s-q0q-t-0-5-p-2-r-1-i-abs-f-df-sum-max}
\end{figure}
\begin{figure}
  \caption{Classifying 10000 chips with 2\% training set size.}
  \label{fig:class10000}
  \centering
\usegraphics{270}{0}{0.9\textwidth}{0.9\textwidth}{1-to-10000-B-s-q0q-t-0-5-p-2-r-1-i-abs-f-df-sum-max}
\end{figure}
Unlike with object identification we use only a subset of positive objects $T$
as training set.  Note that---in the form we present here---we take only
positive objects for training.  Of course we can extend this algorithm to
using a set of positive training objects $T^{1}$ and a set of negative
training objects $T^{0}$ then predict an object under test $x$ to be negative
if $s(x, T^{0})>C^{0}$ where $C^{0}$ is a second cutoff in addition to the
cutoff $C^{1}=C$ we use in Algorithm 3.  Then we would need to extend
Algorithm 4 to find an optimum pair of cutoffs $(C^{1}_{opt},C^{0}_{opt})$
rather than finding just $C_{opt}$ by replacing the one-dimensional base range
or grid by a two-dimensional base range or grid, respectively.

In Figure \ref{fig:class1000} and Figure \ref{fig:class10000} we see plots
showing $s(x,T)$ for 1000 and 10000 objects $x$ where objects are represented
by measurement vectors of chips from real-world chip fabrication.  The
training set size $|T|$ is 2\% of all defective chips here.  The chips have
been reordered so that all defective chips appear on the left and all
non-defective chips on the right. This is just for making the discrepancy of
$s(x,T)$ for defective and non-defective chips $x$ clearly visible.  As can be
seen, defective and non-defective chips are properly separated by their
$s(x,T)$ values and this important property does not depend on the batch
size---1000 or 10000---in this example.

\begin{figure}
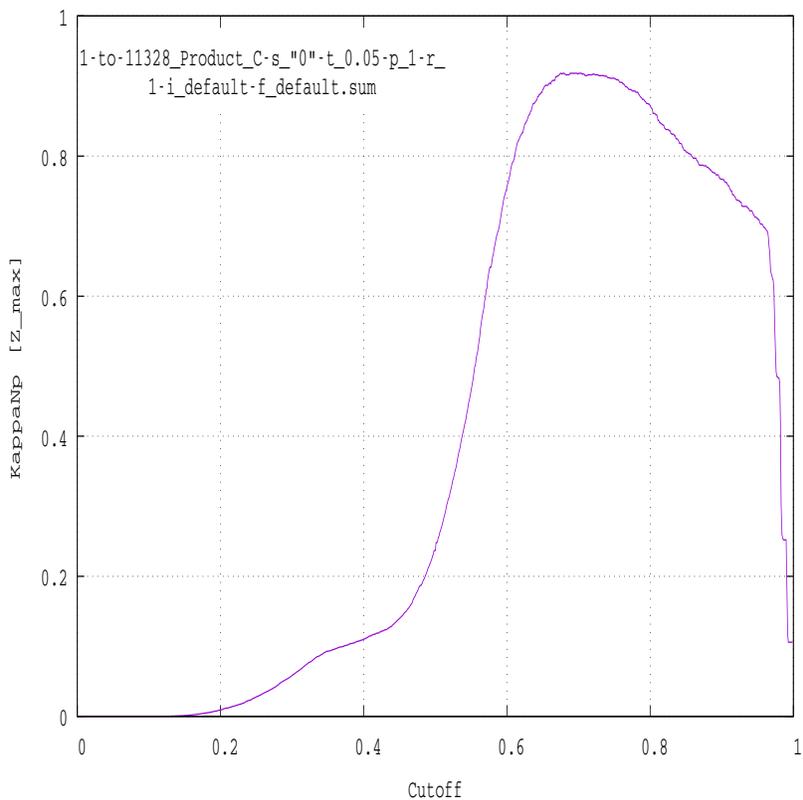

  \caption{Kappa(cutoff) using t-excess thresholding.}
  \label{fig:kappa-cutoff-default}
  \centering
\usegraphics{270}{0}{0.9\textwidth}{0.9\textwidth}{1-to-11328-C-s-q0q-t-0-05-p-1-r-1-i-df-f-df-plot-pick-kp-max}
\end{figure}

\begin{figure}
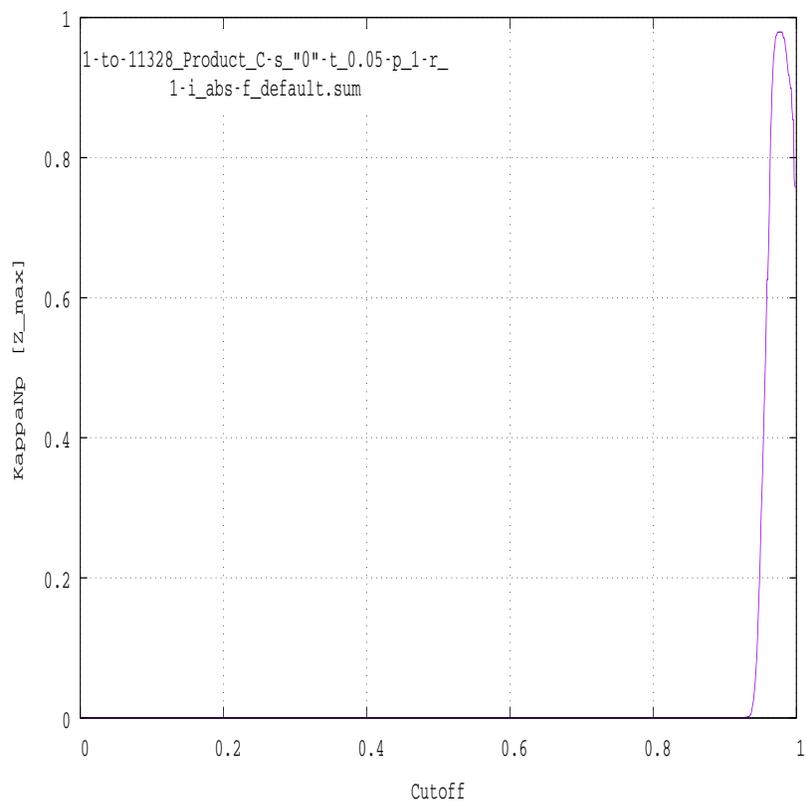

  \caption{Kappa(cutoff) using abs-t-excess thresholding.}
  \label{fig:kappa-cutoff-abs}
  \centering
  \usegraphics{270}{0}{0.9\textwidth}{0.9\textwidth}{1-to-11328-C-s-q0q-t-0-05-p-1-r-1-i-abs-f-df-plot-pick-kp-max}
\end{figure}

Figure \ref{fig:kappa-cutoff-default} and Figure \ref{fig:kappa-cutoff-abs}
show two examples of how the kappa value of predictions changes when the
cutoff varies from 0 to 1.  This is what Algorithm 4 computes if we use
Cohen's kappa for $Q(v,w)$ and use in Algorithm 3 $s(x,T)$ as defined in
subsection \ref{ssec:DenseFormulation}.  We use an equidistant grid
$\mathcal{C}\subseteq[0,1]$ here.  The first time we use thresholding by
t-excess, the second time by abs-t-excess here.  Both graphs have precisely
one optimum $(C_{opt},Q_{opt})$ whereby the hill of Figure
\ref{fig:kappa-cutoff-abs} is sharper and higher than in Figure
\ref{fig:kappa-cutoff-default} which is rather typical with the type of data
we used in our applications.  Figure \ref{fig:kappa-cutoff-default} and Figure
\ref{fig:kappa-cutoff-abs} have been created by the same run with the same
parameters as have been used in creating Figure \ref{fig:class11328-t-excess}
and Figure \ref{fig:class11328-abs-t-excess}.
------------------------------------------------------------------------------
%
%

\section{Experimental Setup}
\label{sec:ExperimentalSetup}
\subsection{The Task}
\label{ssec:TheTask}
In short:
\Bc
\parbox[t]{0.9\textwidth}
{\Em{Given measurements and the defect states for chips of a small training
    set, predict the defect state of the remaining chips of the lot based on
    the measurements only.}
}
\Ec
So we are given measurement data from chip
fabrication.  The data is organized in so-called lots which is a set of wafers
from production. Each wafer carries a fixed number of chips of the same
product type. In what follows we abstract from wafers and consider a lot to be
a set of $m$ chips $x^{i}$ of the same product type.  Each chip $x^{i}$ is
represented by an $n$-vector of measurements $(x^{i}_{1},\ldots,x^{i}_{n})$
and its defect state $p(x^{i})\in\MbB$.  The meaning and ordering of these
measurements is the same for all chips of some lot.  The defect state is one
bit per chip indicating whether the chip is to be considered OK or defective.

One general hypothesis our approach is based on is that defective devices can
be predicted to some extent by looking at combinations of deviations from the
respective component-wise means in measurement data.

Using the notation of subsection \ref{ssec:MachineLearning}, let
$D=\{x^{1},\ldots,x^{m}\}$ be the set of data points.  A positive (negative)
data point is a chip which is to be considered defective (non-defective) and
$D^{1}$ ($D^{0}$) are the sets of defective (non-defective) chips,
respectively.

The research question to be solved is: Find algorithms that are able to learn
the function $f:\MbR^{n}\longrightarrow\MbB$ which assigns the vector of
measurements to the defect state of the respective chip:
\[ (x_{1},\ldots,x_{n})\mapsto p(x)
\]
So, given $(x^{i}_{1},\ldots,x^{i}_{n})$ for a small set of training chips
$x^{i}\in T\subseteq D^{1}$ we want an algorithm that predicts $p(x)$ for
chips $x$ of the same product type with defect state yet unknown, knowing only
the measurement vector $(x_{1},\ldots,x_{n})$ of $x$.  Note that we only need
defect chips for training.

It is easy to see how to extend our algorithm to using non-defective chips for
training, too.  Using the notation of subsection \ref{ssec:Application2},
compute a second indicator
\[ s_{2}(x, T_{2}) = \max_{y\in T_{2}}
             \frac{\sum_{i=1}^{s}\left(1-P_{i}(x)\right)
                     \cdot \left(1-P_{i}(y)\right)}
                  {\sum_{i=1}^{s}\left(1-P_{i}(x)\right)}
\]
where $T_{2}$ is a random subset consisting of negative objects (=
non-defective chips).  This indicator $s_{2}$ expresses maximum similarity of
the object under test $x$ to the objects in $T_{2}$ where the criterion shall
be to \emph{not} exceed the respective threshold now.  Therefore we replace
$P_{i}(x),P_{i}(y)$ of indicator $s$ by $1-P_{i}(x),1-P_{i}(y)$ in $s_{2}$.

In our experiments we considered chips of 3 different product types A, B and C
with the following properties:
\Bc
\begin{tabular}{|l||l|l|l|}
  \hline
  \text{Product} & \#Chips & \#Measurements/chip & continuous only \\
  \hline
  A &  8280 & 385 & yes \\
  B & 11952 & 332 & yes \\
  C & 11328 & 915 & no \\
  \hline
\end{tabular}
\Ec
With product A and B we restricted measurement data to continuous parameters
(currents, voltages etc.) whereas in product C we also used measurements of
discrete parameters (e.g. counts, registers).

It is important that we do not use any meta-knownledge about the
measurements. Neither do we know the types or units nor the meanings of the
measurements.  All we know is their numerical values and that
$(x_{1},\ldots,x_{n})$ results from the same measurements in the same ordering
for all chips $x$ of some lot.
\subsection{The Implementation}
\label{ssec:TheImplementation}
A program for Application 2 as described in subsection \ref{ssec:Application2}
has been realized in a combination of Python 3 (1296 lines) and Bash (112
lines).  More recent versions of this program use the
\Tt{Matplotlib}\cite{Hunter:2007} and \Tt{Numpy} packages.  A minimalist
version has been implemented in only 31 lines (781 bytes) of Python 3 code.
The ``\Tb{for} $x\in D\setminus T$''-loop has been parallelized by using the
\Tt{multiprocessing} package and a speedup of $\sim 3$ could be achieved on a
4-core microprocessor.

A precursor for Application 1 as described in subsection
\ref{ssec:Application1} has been written in Awk, Bash and Prolog.  This
program was generating Prolog code for formalizing an implication system
specified by the thresholding properties of the positive objects---see
subsubsection \ref{sssec:BuildingAndSolving}---and was using Prolog's
unification algorithm for finding solutions.

\section{Results}
\label{sec:Results}
\subsection{Properties}
\label{ssec:Properties}
Algorithms 3 and 4 meet all of the five goals listed in section
\ref{sec:Introduction} and thus have properties which seem to make it
well-suited for deriving the defect state of chips knowing measurement data.

As for the 5th goal---``reducing the number of tests''--- we were successful
even when limiting input data to continuous parameters while leaving out
discrete parameters which may reduce the number of parameters from 900+ to
300-400.  From these we could omit more parameters by limiting input data to
the first measurement step and omit yet many more by applying our method of
dimensional reduction as a preprocessing. More on these topics in subsections
\ref{ssec:PredictingKnowingOnlyPartOfMeasurementData} and
\ref{ssec:DimensionalReduction}.

As for the 4th goal---``ability of coping with large amount of data''--- we
expect it to be possible to handle even much larger lots of data facing the
favorable runtime behaviour of our algorithm.

The hardware demands of all algorithms presented in this paper are low.  The
speed of classifications performed for generating all data mentioned in this
paper ranged from 8738 objects/sec to 415 objects/sec on a Pentium i5-750 (4
$\times$ 2.67 GHz) with a mean of 3068 objects/sec (3rd and 4th goal).
\subsection{Free Parameters}
\label{ssec:FreeParameters}
Using Algorithms 3 and 4 we had three free parameters:
\Bi
\item $p$ (= percentage of training set $T$ from all positive objects)
\item seed (for randomly selecting $T$)
\item threshold $t$
\Ei
  
The meaning of $t$ is explained in subsection \ref{ssec:SettingsOfPractical}.
We were using sets $B_{i}$ defined by the thresholding functions
$P_{i}^{e}(\cdot)$ ($t$-excess) and $P_{i}^{a}(\cdot)$ (abs-$t$-excess)
described there.

The value of $t$ influences the quality of prediction to some extent but it
turned out that one can get good results even with a default $t=0.5$.  We were
setting seed=1 mostly.  In our experiments we were testing a wide range of $p$
from 1 to 50.  Note that---unlike with other algorithms---$p$ refers to a
percentage of the positive objects only.  If, for example, 10\% of all chips
in a lot are defective and we set $p=2$ then Algorithm 1 takes only 0.2\% (not
2\%) of all chips for training, all of which are defective.  Since defective
chips were a small minority in a lot our training sets were frequently much
smaller than with other algorithms applying the ``10:90''- or ``20:80''-rule
in determining the training set size.

In the following tables we will be writing $z_{\max}$ and $z_{\min}$ for two
types of coincidence index, namely $z_{\max}$ for
$s(x, T) = \max\{s(x,y)\mid y\in T\}$ and $z_{\min}$ for
$s(x, T) = \min\{s(x,y)\mid y\in T\}$ using notations of ``\Tb{Task:}'' in
subsection \ref{ssec:DenseFormulation}.
\\~\\
\subsection{Results for Application 2}
\label{ssec:ResultsForApplication2}
The first problem we set out to solve\footnote{Actually the very first problem
  was identification of defective chips---described by Application 1 in
  subsection \ref{ssec:Application1}---the promising results of which were
  encouragement for us to tackle the problem baselying to Application 2.} was
to predict the defect state of chips knowing only a set of measurements as
is described by Application 2 in subsection \ref{ssec:Application2}.
\Bc
\begin{tabular}{|l|l|l|l|l|l||l|l|}
  \hline
  Pro- & \#Chips      & Train & $t$ & Coinc. & Thre-     & Accu- & Kappa\\
  duct & classified   & set   &     & index  & sholding  & racy  &      \\
  \hline
  A & 5000 & 35\% & 0.48 & $z_{\max}$ & t-excess & 0.943 & 0.484 \\
  A & 5000 & 35\% & 0.94 & $z_{\max}$ & abs-t-excess & 0.954 & 0.633 \\
  \hline
  B & 5000 & 2\% & 0.5 & $z_{\max}$ & t-excess & 0.967 & 0.788 \\
  B & 5000 & 2\% & 0.5 & $z_{\max}$ & abs-t-excess & 0.990 & 0.938 \\
  \hline
  B & 10000 & 15\% & 0.5 & $z_{\max}$ & t-excess & 0.979 & 0.824 \\
  B & 10000 &  1\% & 0.5 & $z_{\max}$ & abs-t-excess & 0.990 & 0.929 \\
  \hline
  C & 11328 &  1\% & 0.05 & $z_{\max}$ & t-excess & 0.992 & 0.918 \\
  C & 11328 &  1\% & 0.05 & $z_{\max}$ & abs-t-excess & 0.998 & 0.979 \\
  \hline
\end{tabular}
\Ec
As can be seen, thresholding by 1 iff $|x_{i}|>t$ (abs-t-excess in the above
table, $P_{i}^{a}(x)$ in subsection \ref{ssec:SettingsOfPractical}) typically
leads to better results than just $x_{i}>t$ (t-excess, $P_{i}^{e}(x)$).  This
is not surprising because the latter does not catch conditions $x_{i}<-t$
(i.e. measurements being notably lower than the mean value) whereas the former
does.

\begin{figure}
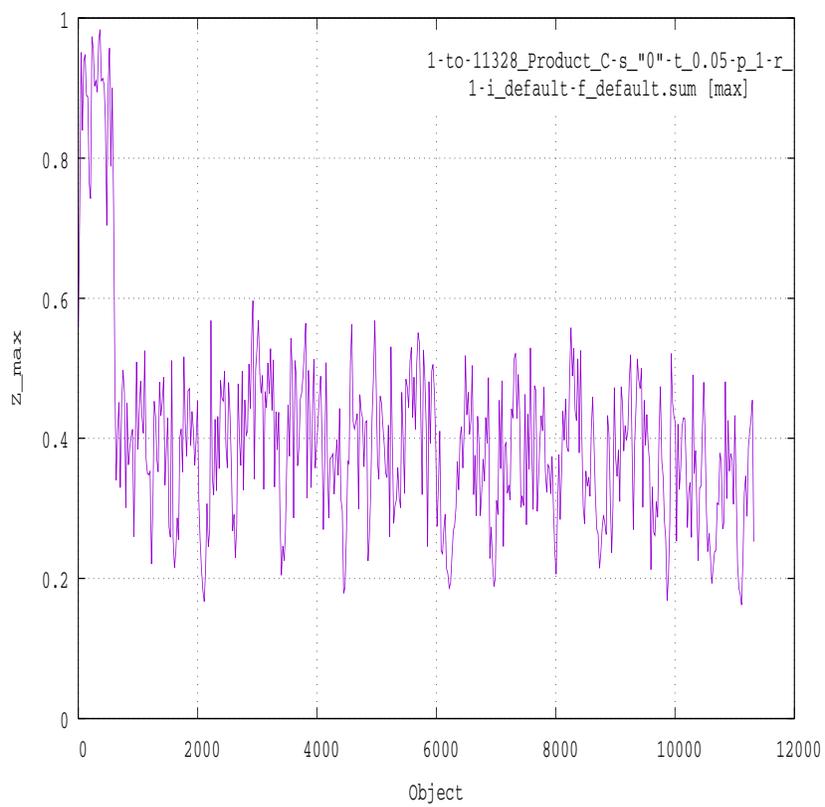

  \caption{$z_{\max}$ using t-excess for thresholding.}
  \label{fig:class11328-t-excess}
  \centering
  \usegraphics{270}{0}{0.9\textwidth}{0.9\textwidth}{1-to-11328-C-s-q0q-t-0-05-p-1-r-1-i-df-f-df-sum-max}
\end{figure}

The next two plots---Figure \ref{fig:class11328-t-excess} and Figure
\ref{fig:class11328-abs-t-excess}---show $z_{\max}$ at classifying a
complete lot of 11328 chips of product C in one run with the aforementioned
two different thresholding functions t-excess ($x_{i}>t$) and abs-t-excess
($|x_{i}|>t$).
\begin{figure}
  \caption{$z_{\max}$ using abs-t-excess for thresholding.}
  \label{fig:class11328-abs-t-excess}
  \centering
\usegraphics{270}{0}{0.9\textwidth}{0.9\textwidth}{1-to-11328-C-s-q0q-t-0-05-p-1-r-1-i-abs-f-df-sum-max}
\end{figure}
Both plots show how easy it is to place a horizontal cutoff line in order to
cleanly separate the defective chips from the non-defective chips by the
discrepancies of their $z_{max}$ values.  As the corresponding two lines in
the first table in section \ref{ssec:ResultsForApplication2} show,
abs-t-excess is slightly superior to t-excess here with a kappa value of 0.979
vs. 0.918.\footnote{Both plots look like it were easily possible to find a
  cutoff value which would separate the defective chips from the rest with an
  accuracy of 100\% and kappa value of 1.0. In truth there are some few peaks
  in these graphs of $z_{max}$ which are fine enough for being invisible in
  lack of unlimited rendering resolution but which push down both accuracy and
  kappa slightly below the optimum of 1.0.}

\begin{figure}
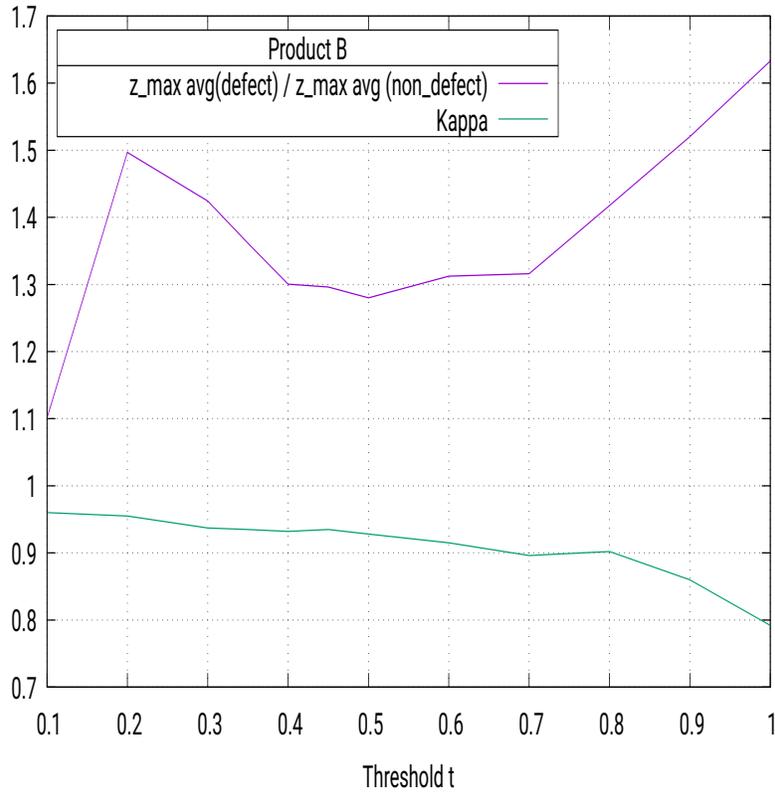

  \caption{Quotient of $z_{\max}$ averages and kappa.}
  \label{fig:quot-avg-and-kappa}
  \centering
  \usegraphics{270}{0}{0.9\textwidth}{0.9\textwidth}{param-study1a}
\end{figure}

\begin{figure}
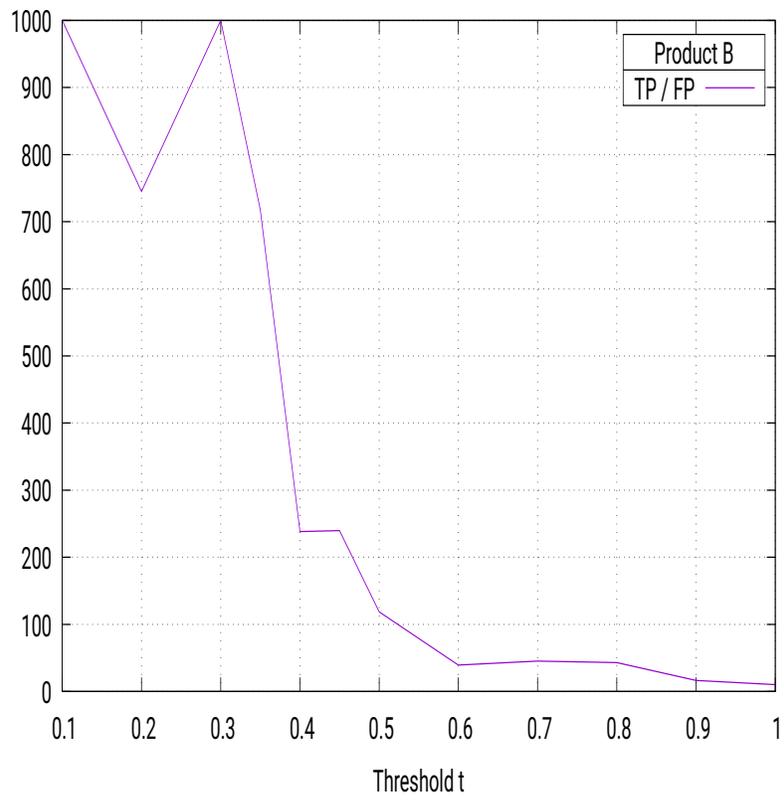

  \caption{$\frac{\Mr{TP}}{\Mr{FP}}$ over threshold $t$.}
  \label{fig:quot-tp-fp}
  \centering
  \usegraphics{270}{0}{0.9\textwidth}{0.9\textwidth}{param-study1b}
\end{figure}

Figure \ref{fig:quot-avg-and-kappa} is an example of how kappa and the
quotient of averages
\[
   Q_{\text{avg,$z_{max}$}} = \frac{\text{avg $z_{max}$(defective chips)}}
   {\text{avg $z_{max}$(non-defective chips)}}
\]
depend on threshold $t$.  The data stems from classifying 10000 chips of
product B in one run with 2\% training set size using abs-t-excess
thresholding with threshold $t$ varying from 0.1 to 1. Even though intuition
tells that higher $Q_{\text{avg,$z_{max}$}}$ values tend to be associated with
higher kappa values there is not a direct relation of high
$Q_{\text{avg,$z_{max}$}}$ values and high kappa.  One reason for this lies in
the fact that in order to find a good $z_{max}$ cutoff for separating
defective chips from non-defective ones other properties like variance and
regularity/roughness of the $z_{max}$ graph---see Figure
\ref{fig:class11328-t-excess} and Figure
\ref{fig:class11328-abs-t-excess}---play important roles.  Figure
\ref{fig:quot-avg-and-kappa} also shows the excellent kappa levels---0.96 in
this setting---that can be reached by our algorithms and the relative
independence of kappa from threshold $t$ here: kappa $>0.78$ irregardless of
the choice of $t\in[0,1]$.
Figure \ref{fig:quot-tp-fp} stems from the same data and parameters as Figure
\ref{fig:quot-avg-and-kappa} and shows an example of how the
true-positive-false-positive quotient depends on threshold $t$.  A high
$\frac{\Mr{TP}}{\Mr{FP}}$ value is an important performance criterion of
methods for defect prediction in semiconductor fabrication because high
$\frac{\Mr{TP}}{\Mr{FP}}$ values directly translate into reducing costs by
avoiding to scrap non-defective devices based on false predictions.  As Figure
\ref{fig:quot-tp-fp} shows, by applying our algorithmic method we risk
scrapping only one functioning device per 1000 defective devices predicted
correctly which is an excellent performance facing that minimum demands may be
as low as 2.
\subsubsection{Predicting Specific Types of Failure}
\label{ssec:PredictingSpecificTypesOfFailure}
In quality control of chip fabrication it is informative to know not only that
a chip will be defective---and thus should not be bonded, boxed and
shipped---but also the specific type of defect.  Defective chips can be
classified by assigning those chips to the same soft bin (SBIN) which suffer
from the same type of defect.

Predicting whether the type of defect of some chip under test belongs to a
certain SBIN $S$ can be formulated as a variant of the aforementioned problem
of predicting the defect state of chips.  For doing this, we just substitute
the target function $p:D\rightarrow\MbB$ we used for defect state prediction,
\[
  \text{$p(x)=1$ iff data point $x$ belongs to a defective chip,}
\]
by:
\[
  \text{$p(x)=1$ iff data point $x$ is assigned to $S$.}
\]
\Bc
\begin{tabular}{|l|l|l|l|l|l|l||l|l|}
  \hline
  Pro- & SBIN & Chips      & Train & $t$ & Coinc.   & Thre-     & Accu-
& Kappa\\
  duct &      & classi-      & set   &     & index  & sholding  & racy
&      \\
       &      & fied         &       &     &        &         &     
&      \\
  \hline
  B & "6" & 11952 & 30\% & 0.05 & $z_{\max}$ & t-excess & 0.999 & 0.749 \\
  B & "7" & 11952 & 50\% & 0.05 & $z_{\min}$ & t-excess & 1.000 & 0.895 \\
  \hline
\end{tabular}
\Ec
We have chosen SBINs "6" and "7" because these two were the SBINs most
frequently occurring in the baselying data material.  Though there were very
few chips with these defect types---only 50 and 40 out of 11952 chips, which
is {4.2\textperthousand} and {3.3\textperthousand}, respectively.  So even in
these two cases there were very few positive objects to be used for training.
Because it is not possible to learn if the sample frequency is too low we
suspect that if we had more samples of defective chips with a specific defect
type for disposal then other SBINs could be predicted, too.
\subsubsection{Detecting Iris Type}
\label{ssec:DetectingIrisType}
In order to know whether our algorithm can learn functions from an application
range different than detecting chip defects we applied it to the classic iris
flower data set\cite{Fisher:1936} in the corrected form of \cite{Dua:2019}
(samples \#35 and \#38).
\Bc
\begin{tabular}{|l|l|l|l|l||l|l|}
  \hline
  Iris & Train & $t$ & Coinc. & Thre-     & Accu-
& Kappa\\
  type & set   &     & index  & sholding  & racy
&      \\
  \hline
  setosa & 20\% & 0.1 & $z_{\max}$ & t-excess & 0.971 & 0.928 \\
  versicolor & 30\% & 1.0 & $z_{\max}$ & abs-t-excess & 0.852 & 0.557 \\
  virginica & 30\% & 0.9 & $z_{\max}$ & t-excess & 0.919 & 0.797 \\
  \hline
\end{tabular}
\Ec
As one can see, setosa and virginica can be detected quite successfully
whereas detecting versicolor seems harder to our algorithm.  This is in
accordance with the---both expected and observed---strengths of our algorithm:
Looking at visualizations of the iris flower data set one sees that sepal and
petal length and width, resp., of the versicolor species are in a medium range
whereas sepal and petal length and width, resp., of setosa and virginica are
on the lower and upper ends of the value ranges.  Since our algorithms are
made for identifying and detecting defective chips---where certain
combinations of exceedingly low or exceedingly high values frequently
indicate a defective chip---it is no surprise that detecting setosa and
virginica is rather easy to our algorithm whereas detecting versicolor is
harder.
\subsubsection{Prototypical Positive Objects}
\label{ssec:PrototypicalPositiveObjects}
In order to gain insights into why defective chips can be detected by looking
for maximum similarity regarding $s(x,T)$---see subsection
\ref{ssec:Application2}---we created histograms for answering the following
questions:
\\~\\
\Tb{Given}: chip $y\in T$
\\~\\
\Tb{Question}: how big is $\big|\{ x\in D^{1}\mid s(x,y)=s(x,T)\}\big|$?
\\~\\
We will call the latter number $n(y)$ and define
\[ H(y):=\frac{n(y)}{|D^{1}\setminus T|}
\]
In words, $H(y)$ is the amount of those defective chips which have maximum
similarity to $y\in T$ in all defective chips outside the training set.
Remember that we only classify chips outside the training set $T$ because we
do not consider classifying chips of the training set an interesting goal in
Application 2.

An interesting observation in most of our experiments was that in most
settings $H(y)$ was relatively large for a small number of chips $y\in T$ with
a steep decrease towards those $y\in T$ with smaller $H(y)$.
\\~\\
\Tb{Example} of a full histogram (product B, 5000 chips to be classified, 13\%
training set):
\Bc
\begin{tabular}{|r|r|r|}
  \hline
  \multicolumn{1}{|c|}{$y$ (= Train} & \multicolumn{1}{|c|}{$H(y)$}
    & \multicolumn{1}{|c|}{$n(y)$} \\
  \multicolumn{1}{|c|}{object ID)} & \multicolumn{1}{|c|}{\%}
    & \multicolumn{1}{|c|}{(\#chips)} \\
  \hline
        3762 & 54.39 &  223 \\
        8407 & 8.54 &   35 \\
        8342 & 8.29 &   34 \\
       14361 & 5.85 &   24 \\
        3461 & 3.66 &   15 \\
        7185 & 3.41 &   14 \\
          51 & 3.41 &   14 \\
          33 & 2.93 &   12 \\
        9650 & 2.20 &    9 \\
        2875 & 2.20 &    9 \\
        4856 & 0.98 &    4 \\
        7007 & 0.98 &    4 \\
         159 & 0.73 &    3 \\
       11586 & 0.49 &    2 \\
        3693 & 0.49 &    2 \\
        1168 & 0.49 &    2 \\
        6927 & 0.49 &    2 \\
         307 & 0.24 &    1 \\
        7963 & 0.24 &    1 \\
  \hline
\end{tabular}
\Ec
As can be seen easily, almost all defective chips outside the training set are
most similar---regarding $s(x,T)$, see subsection \ref{ssec:Application2}---to
only a small subset of the training samples. In this example, $\sim$95\% of
all defective chips (410 in this example) are most similar to only 10 training
samples, namely the 10 topmost chips in the table above.  Chips who occur near
the top of such a histogram possess specific properties that allow for
detecting a multitude of defective chips.  We call these chips
\Em{prototypical defective chips}.

In the following table we show $n(y)$ and $H(y)$ for the topmost 3 training
chips $y_{1},y_{2},y_{3}\in T$ of the respective histogram.  The parameters
belonging to the lines of this table are the same as in the first table of
subsection \ref{ssec:ResultsForApplication2}.
\Bc
\begin{tabular}{|l|l|l|l||l|l|}
  \hline
  Pro- & \#Chips      & Train & Thre-    & Accu- & Kappa  \\
  duct & classified   & set   & sholding & racy  &        \\
  \hline
  A & 5000 & 35\% & t & 0.943 & 0.484 \\
  A & 5000 & 35\% & abs-t & 0.954 & 0.633 \\
  \hline
  B & 5000 & 2\% & t & 0.967 & 0.788  \\
  B & 5000 & 2\% & abs-t & 0.990 & 0.938 \\
  \hline
  B & 10000 & 15\% & t & 0.979 & 0.824  \\
  B & 10000 &  1\% & abs-t & 0.990 & 0.929 \\
  \hline
  C & 11328 &  1\% & t & 0.992 & 0.918  \\
  C & 11328 &  1\% & abs-t & 0.998 & 0.979  \\
  \hline
\end{tabular}
\Ec
\Bc
\begin{tabular}{|l|l|l|l||l|l|l|}
  \hline
  Pro- & \#Chips      & Train & Thr. 
     & $H(y_{1})$\% & $H(y_{2})$\% & $H(y_{3})$\% \\
  duct & classified   & set   &
     & $\big(n(y_{1})\big)$ & $\big(n(y_{2})\big)$ & $\big(n(y_{3})\big)$ \\
  \hline
  A & 5000 & 35\% & t & 3\% (12) & 3\% (12) & 3\% (12) \\
  A & 5000 & 35\% & abs-t & 11\% (38) & 5\% (18) & 5\% (18) \\
  \hline
  B & 5000 & 2\% & t & 27\% (126) & 17\% (80) & 13\% (59) \\
  B & 5000 & 2\% & abs-t & 33\% (151) & 32\% (147) & 16\% (72) \\
  \hline
  B & 10000 & 15\% & t & 21\% (146) & 7\% (47) & 5\% (35) \\
  B & 10000 &  1\% & abs-t & 40\% (322) & 38\% (308) & 15\% (122) \\
  \hline
  C & 11328 &  1\% & t & 34\% (204) & 28\% (171) & 19\% (116) \\
  C & 11328 &  1\% & abs-t & 38\% (230) & 37\% (223) & 12\% (71) \\
  \hline
\end{tabular}
\Ec
{\small (Thr. = Thresholding)}
\\~\\
As in the full histogram example, we see that most of the defective chips to
be classified have maximum similarity to only a small set of $y\in T$.  It may
be worthwhile to investigate properties of prototypical defective chips more
in-depth for they seem to embody a kind of ``blueprint'' for whole classes of
defective chips. So modifying fabrication with the goal of avoiding
prototypical defective chips might eliminate whole classes of defective
devices.

This observation also held true when applying our algorithm to the iris flower
classification problem.  In these histograms we found one sample $y$ with
$H(y)=90\%$, $H(y)\sim 49\%$ and $H(y)\sim 91\%$ when attempting to detect
setosa, versicolor and virginica, resp. 

Returning to chip measurement data---as can be seen from the above table,
accuracy and kappa are worse with product A than with products B and C and the
steepness of $H(y)$ for product A is low if compared to products B and C.
Whether those chip measurement data sets which are generally easier to our
algorithm are associated with higher Gini coefficients of $H(y)$ may be
subject of further research.\footnote{The Gini coefficient $G$ measures a
  degree of unequality in a distribution\cite{Gini:1912}.
  If $y_{1},\ldots,y_{n}$ are
  non-negative values of $n$ objects then
  $G=\frac{\sum_{i=1}^{n}\sum_{j=1}^{n}|y_{i}-y_{j}|}
  {2 n\cdot\sum_{i=1}^{n}y_{i}}\in[0,1]$.}
The following table shows the confusion matrices belonging to the lines of the
first table of subsection \ref{ssec:ResultsForApplication2}.
\Bc
\begin{tabular}{|l||l|l||l|l|l|l||l|}
  \hline
  Pro- & Accu- & Kappa & TP & FP & TN & FN & \underline{TP} \\
  duct & racy  &       &    &    &    &    &  FP \\
  \hline
  A & 0.943 & 0.484      & 118 & 34 & 4418 & 241 & 3.5 \\
  A & 0.954 & 0.633  & 153 & 17 & 4435 & 206 & 9.0 \\
  \hline
  B & 0.967 & 0.788 & 331 & 34 & 4499 & 131 & 9.7 \\
  B & 0.990 & 0.938   & 417 & 5 & 4528 & 45 & 83.4 \\
  \hline
  B & 0.979 & 0.824     & 535 & 47 & 9129 & 164 & 11.4 \\
  B & 0.990 & 0.929 & 719 & 4 & 9172 & 97 & 179.8 \\
  \hline
  C & 0.992 & 0.918     & 524 & 4 & 10709 & 84 & 131.0 \\
  C & 0.998 & 0.979 & 584 & 0 & 10713 & 24 & $\infty$ \\
  \hline
\end{tabular}
\Ec
Facing that a minimum requirement for considering a defect chip detection
method to be feasible might be $\frac{TP}{FP}>1$---that is, allowing for
scrapping at most one intact chip for every defective chip being properly
detected in the long run---the confusion matrices in the above table are very
good to excellent.  It is even possible to reach $FP=0$ which implies
$\frac{TP}{FP}=\infty$ with product C.
\subsubsection{Predicting Defect States Knowing Only Part of Measurement Data}
\label{ssec:PredictingKnowingOnlyPartOfMeasurementData}
In this subsection we describe our solution to a set of important problems
circling around saving costs by omitting measurements while computing still
accurate and meaningful predictions.  Differing relevant measurements from
those which are irrelevant for keeping a certain quality level and thus can be
omitted is generally a problem of high importance in fabrication and quality
assurance.  On the one side measurements in fabrication are costly and take
time so it is desirable to reduce the amount of measurements carried out to a
minimum.  On the other side sophisticated measurements enable detecting
defects early and the earlier they are detected the better and---mostly---the
cheaper.  In addition, the time and space complexity of typical classification
algorithms depends at least linearly on the number of features which may be an
extra motivation for example in environments with real-time demands.

The measurement data we had at our disposal consisted in parameter packs
---so-called measurement steps ---carrying names like T1, T2, Y5.  Each
measurement belongs to a certain measurement step.  Measurements of different
measurement steps differ in the step of fabrication they were taken at and their type.
So T1 measurements are carried out in a measurement step of a fabrication step
earlier than T2 measurements.

To be more specific, the problem we turn to now can be formulated like this.
\Bc
\Em{Is it possible to predict the defect state that would appear after
  measurement step T2 if all we know is measurement data of measurement step T1?}
\Ec
This problem is harder: The classifier has to predict the T2 defect state
without knowing any T2 measurements.

So in case of product B the first 77 (T1) of 332 (T1+T2) measurements per chip
are given as input to the classification algorithm, and in case of product C,
only the first 429 (T1) of 915 (T1+T2) measurements per chip are given.

Our results show that our algorithmic methods are a means to solve this harder
problem, too, with good to very good kappa values and excellent accuracy.
\Bc
\begin{tabular}{|l|l|l|l|l|l||l|l|}
  \hline
  Pro- & \#Chips      & Train & $t$ & Coinc. & Thre- & Accu-
& Kappa\\
  duct & classified   & set   &     & index  & sholding  & racy
&      \\
  \hline
  B & 10000 & 10\% & 0.55 & $z_{\max}$ & abs-t-excess & 0.970 & 0.772 \\
  B & \quad$\shortparallel$ & \ \;$\shortparallel$ & \ \;$\shortparallel$ & $z_{\min}$ & \qquad$\shortparallel$ & 0.968 & 0.713 \\
  C & 11328 & 1\% & 0.05 & $z_{\max}$ & \qquad$\shortparallel$ & 0.979 & 0.760 \\
  C & \quad$\shortparallel$ & \ \;$\shortparallel$ & \ \;$\shortparallel$ & $z_{\min}$ & \qquad$\shortparallel$ & 0.989 & 0.885 \\
  \hline
\end{tabular}
\Ec

Figure \ref{fig:use-s1-tell-s2-prod-b} and Figure
\ref{fig:use-s1-tell-s2-prod-c} show how $z_{\max}$ or $z_{\min}$---as
computed by Algorithm 3---can be used to classify in one run the T2 defect
state of 10000 and 11328 chips knowing only T1 measurements.  Figure
\ref{fig:kappa-cutoff-use-s1-tell-s2-prod-b} and Figure
\ref{fig:kappa-cutoff-use-s1-tell-s2-prod-c} display the graph $\Gamma_{Q}$
belonging to Figure \ref{fig:use-s1-tell-s2-prod-b} and Figure
\ref{fig:use-s1-tell-s2-prod-c}.
\begin{figure}
  \caption{$z_{\max}$ predicting T2 defect state knowing only T1 measurements.}
  \label{fig:use-s1-tell-s2-prod-b}
  \centering
  \usegraphics{270}{0}{0.9\textwidth}{0.9\textwidth}{1-to-10000-B-useS1-tellS2-s-q0q-t-0-55-p-10-r-1-i-abs-f-df-sum-max}
\end{figure}
\begin{figure}
  \caption{Kappa(cutoff) of the foregoing figure.}
  \label{fig:kappa-cutoff-use-s1-tell-s2-prod-b}
  \centering
  \usegraphics{270}{0}{0.9\textwidth}{0.9\textwidth}{1-to-10000-B-useS1-tellS2-s-q0q-t-0-55-p-10-r-1-i-abs-f-df-plot-pick-kp-max}
\end{figure}
\begin{figure}
  \caption{$z_{\min}$ predicting T2 defect state knowing only T1 measurements.}
  \label{fig:use-s1-tell-s2-prod-c}
  \centering
  \usegraphics{270}{0}{0.9\textwidth}{0.9\textwidth}{1-to-11328-C-useS1-tellS2-s-q0q-t-0-05-p-1-r-1-i-abs-f-df-sum-min}
\end{figure}
\begin{figure}
  \caption{Kappa(cutoff) of the foregoing figure.}
  \label{fig:kappa-cutoff-use-s1-tell-s2-prod-c}
  \centering
  \usegraphics{270}{0}{0.9\textwidth}{0.9\textwidth}{1-to-11328-C-useS1-tellS2-s-q0q-t-0-05-p-1-r-1-i-abs-f-df-plot-pick-kp-min}
\end{figure}

\Tb{Prototypical defective chips.}  In the following two tables we list the
top lines of histograms counting for each training chip $y\in T$ how many
defective chips $x\in D^{1}\setminus T$ to be classified have maximum $s(x,t)$
for $t=y$.  See \ref{ssec:PrototypicalPositiveObjects} for definitions of
$n(y)$ and $H(y)$.

\Tb{Top of histograms} for product B (left table) and product C (right table):
\Bc
\begin{tabular}[t]{|r|r|r|}
  \hline
  \multicolumn{1}{|c|}{$y$ (= Train} & \multicolumn{1}{|c|}{$H(y)$}
    & \multicolumn{1}{|c|}{$n(y)$} \\
  \multicolumn{1}{|c|}{object ID)} & \multicolumn{1}{|c|}{\%}
    & \multicolumn{1}{|c|}{(\#chips)} \\
  \hline
         148 & 23.69  &  176 \\
        7032 & 18.44  &  137 \\
          38 & 16.02  &  119 \\
         195 & 7.00   &   52 \\
        1019 & 5.11   &   38 \\
        6365 & 4.98   &   37 \\
       10020 & 4.85   &   36 \\
          60 & 3.36   &   25 \\
        8526 & 3.36   &   25 \\
        2504 & 2.69   &   20 \\
        9984 & 2.15   &   16 \\
       11464 & 1.88   &   14 \\
        1718 & 1.75   &   13 \\
         977 & 1.35   &   10 \\
  \hline
\end{tabular}
\qquad
\begin{tabular}[t]{|r|r|r|}
  \hline
  \multicolumn{1}{|c|}{$y$ (= Train} & \multicolumn{1}{|c|}{$H(y)$}
    & \multicolumn{1}{|c|}{$n(y)$} \\
  \multicolumn{1}{|c|}{object ID)} & \multicolumn{1}{|c|}{\%}
    & \multicolumn{1}{|c|}{(\#chips)} \\
  \hline
         235 & 37.50 &  228 \\
       10392 & 36.68 &  223 \\
        6030 & 11.68 &   71 \\
        9193 & 10.20 &   62 \\
        6194 &  3.62 &   22 \\
  \hline
\end{tabular}
\Ec

For both products we see again the phenomenon of prototypical defective chips
described in \ref{ssec:PrototypicalPositiveObjects}: Relatively few chips $y$
have maximum $s(x,y)$ with the vast majority of defective chips.

\Tb{An easier variant.} A variant of minor practical importance is the task of
learning the T1 defect state knowing T1 and T2 measurements.  We have tackled
this task using our algorithms and could reach an accuracy of 100.0\% and
kappa 0.997 with only 2 out of 10000 chips having been classified wrongly at
classifying 10000 chips of product B using $z_{max}$, 40\% training set,
threshold 0.04.  When using $z_{min}$ results are only slightly worse with
accuracy 99.7\% and kappa 0.961.
\subsubsection{Dimensional Reduction}
\label{ssec:DimensionalReduction}
As explained in subsection
\ref{ssec:PredictingKnowingOnlyPartOfMeasurementData}, reducing the number of
measurements without experiencing an intolerable reduction in failure
detection quality is a worthwhile task.  In subsection
\ref{ssec:PredictingKnowingOnlyPartOfMeasurementData} we investigated how all
measurements of a complete measurement step can be omitted.

To go further we developed a method for reducing the number of measurements by
omitting features with specific properties as a preprocessing step.  This
method has one free parameter we call \Em{sharpness} by which we can tune the
amount of reduction.

\Tb{The algorithm.}  The set of input data points $D$ must be given as lines
of a $m\times n$-matrix, auto-scaled by column.  Algorithm 2 can be used for
this.  In what follows we use ``$\sqcup$'' as denotation for disjunctive
decomposition of a set.  $\{1,\ldots,m\}=I^{0}\sqcup I^{1}$ is the index
decomposition induced by splitting $D$ into negative and positive objects,
$D=D^{0}\sqcup D^{1}$.
\\
\begin{algorithm}[H]
  \KwInput{$D=\{x_{1},\ldots,x_{m}\}\subseteqq \MbR^{n}$
    ($m$ line vectors) auto-scaled by column,
    $\{1,\ldots,m\}=I^{0}\sqcup I^{1}$}
  \KwInput{$\Mr{sharpness}\in\MbN_{\geqq0}$}
  \KwOutput{$E\in \MbR^{m\times n^{\ast}}$, $n^{\ast}$}
\caption{dim-reduce($D$, $\Mr{sharpness}$)}
\For{$i\in I^{1}$}
{
  m := $\max\{|x_{ij}|:j\in\{1,\ldots,n\}\}$\\
  $\Mr{MaxIndices}_{i} := \{j\in\{1,\ldots,n\}:|x_{ij}|=m\}$
}
\For{$j=1\ldots n$}
{
  $\Mr{NumOccu}_{j} := \mathlarger{\sum\limits_{\substack{i=1,\ldots,m\\ i\in
        I^{1}}}}
  |\Mr{MaxIndices}_{i}\cap\{j\}|$ \\
  \tcp{count how many positive objects $x_{i}\in I^{1}$ have maximum}
  \tcp{absolute value in the $j$-th component of the auto-scaled data}
}
$k := 0$ \\
\For{$j=1\ldots n$}
{
  \If{$|\Mr{NumOccu}_{j}|\geqq\Mr{sharpness}$}
  {
    $k := k + 1$ \\
    \For{$i=1\ldots m$}
    {
      $e_{i,k} := x_{i,j}$
    }
    \tcp{copy column $j$ from $D$ to $E$}
  }
}
$n^{\ast} := k$
\end{algorithm}
Time complexity is linear in the size $m\times n$ of the input matrix if we
neglect polylogarithmic factors.

The following table shows some results of classifying 11328 chips of product C
in one run with parameters known from further above: 1\% training set size,
abs-t-excess thresholding, goal: predicting defect state of measurement step T2
knowing only measurements of T1.  The number of features before reduction is
429 (= total number of measurements per chip in T1) whereas the number of
features after dimensional reduction is shown under \#feat.~omitted.
Sharpness 0 in the first line of the table means no dimensional reduction i.e.
all 429 features are given to the classification algorithm.
\Bc
\begin{tabular}{|r|r|r|r|r|r|}
  \hline
  \multicolumn{1}{|c|}{Sharpness} & \multicolumn{1}{|c|}{\#feat.}
    & \multicolumn{1}{|c|}{\%feat.}
    & \multicolumn{1}{|c|}{Accuracy} & \multicolumn{1}{|c|}{Kappa}
    & \multicolumn{1}{|c|}{\underline{TP}} \\
  \multicolumn{1}{|c|}{of reduct.} & \multicolumn{1}{|c|}{omitted}
    & \multicolumn{1}{|c|}{omitted}
    & \multicolumn{1}{|c|}{[$z_{min}$]} & \multicolumn{1}{|c|}{[$z_{min}$]}
    & \multicolumn{1}{|c|}{FP} \\
  \hline
       0 &   0 &    0\% & 0.989 & 0.885 & 33.3 \\
       1 & 283 & 66.0\% & 0.974 & 0.664 & $+\infty$ \\
       2 & 346 & 80.7\% & 0.973 & 0.656 & 34.4 \\
       3 & 379 & 88.3\% & 0.946 & 0.194 & 0.234 \\
       4 & 387 & 90.2\% & 0.946 & 0.188 & 0.201 \\
  \hline
\end{tabular}
\Ec

\Tb{Interpretation:} By performing our dimension reduction with sharpness 1 we can
omit 66\% of measurements and still predict T2 defect state with accuracy
97.4\%, kappa 0.664.  With sharpness 2 we can omit 80.7\% of measurements and
reach accuracy 97.3\%, kappa 0.656 which is only marginally worse.  With
sharpness $\geqq3$ the quality of prediction---clearly visible by looking at
kappa and $\frac{\Mr{TP}}{\Mr{FP}}$---declines rapidly so that the
recommendation would be to not go farther than sharpness = 2 and thus save
80.7\% of measurements.

The important true-positive-false-positive quotients are impressively high
when using sharpness $<3$.



\section{Conclusion}
\label{sec:Conclusion}
We have presented a fast generic algorithm and a method for dimensional
reduction for detecting abnormal objects in high-dimensional data which has
been applied to data from wafer fabrication in order to predict defect states
of tens of thousands of chips of several products based on measurements or
part of measurements successfully.

When compared with the optimization principle of typical neural-net based
algorithms, we are looking for the optimal rater of a very limited set of
possible raters---given by the constants computed in the training step and the
structural simplicity of the operations graph and flow-chart of our algorithms
with only one or two undetermined variables to optimize over---instead of
looking for the optimal rater in a high-dimensional set of possible raters
with many undetermined variables to optimize over.

The question whether these methods can be applied in situations where the
connections of measurement data and defect state are of a more indirect nature
like when trying to predict backend defects knowing frontend measurement data
is subject of ongoing research.

\listoffigures
\bibliography{defect-detection}{}
\bibliographystyle{plain}
\end{document}